\documentclass[10pt,twocolumn,letterpaper]{article}

\usepackage{iccv}              
\usepackage{multirow}   
\usepackage{url}
\definecolor{iccvblue}{rgb}{0.21,0.49,0.74}
\usepackage[pagebackref,breaklinks,colorlinks,allcolors=iccvblue]{hyperref}

\def\paperID{7892} 
\def\confName{ICCV}
\def\confYear{2025}

\title{A Theory-Inspired Framework for Few-Shot Cross-Modal Sketch Person Re-Identification}

\author{
	Yunpeng Gong$^{1}$ \quad
	Yongjie Hou$^{2}$ \quad
	Jiangming Shi$^{3}$ \quad
	Kim Long Diep$^{1}$ \quad
	Min Jiang$^{1}$\thanks{Corresponding author: Min Jiang(minjiang@xmu.edu.cn). Min Jiang and Yunpeng Gong(fmonkey625@gmail.com) are with the Department of Artificial
		Intelligence, Key Laboratory of Multimedia Trusted Perception and Efficient Computing, Ministry of Education of China, School of Informatics, Key Laboratory of Digital Protection and Intelligent Processing of Intangible CulturalHeritage of Fujian and Taiwan, Ministry of Culture and Tourism, Xiamen University, Xiamen 361005, Fujian, P.R. China.} \\
	$^{1}$School of Informatics, Xiamen University \\
	$^{2}$School of Electronic Science and Engineering, Xiamen University \\
	$^{3}$Institute of Artificial Intelligence, Xiamen University 
}

\begin{document}
\maketitle

\begin{abstract}
	Sketch-based person re-identification aims to match hand-drawn sketches with RGB surveillance images, but remains challenging due to severe modality gaps and limited labeled data. To address this, we propose KTCAA, a theoretically inspired framework for few-shot cross-modal generalization. Drawing on generalization bounds, we identify two key factors affecting target risk: (1) domain discrepancy, reflecting the alignment difficulty between source and target distributions; and (2) perturbation invariance, measuring the model’s robustness to modality shifts. Accordingly, we design: (1) Alignment Augmentation (AA), which applies localized sketch-style transformations to simulate target distributions and guide progressive alignment; and (2) Knowledge Transfer Catalyst (KTC), which enhances perturbation invariance by introducing worst-case modality perturbations and enforcing consistency. These modules are jointly optimized within a meta-learning paradigm that transfers alignment knowledge from data-abundant RGB domains to sketch scenarios. Experiments on multiple benchmarks show that KTCAA achieves state-of-the-art performance, particularly under data-scarce conditions. The code will be available at \url{https://github.com/finger-monkey/REID_KTCAA}.
\end{abstract}


\section{Introduction}

Person re-Identification (ReID)\cite{zhong2019invariance,zhong2017re,liu2020hierarchical,sun2024alice,wang2024semi,gong2022person,gong2024beyond} aims to match individuals across different images or camera views, playing a key role in surveillance and law enforcement. However, in situations where photographic evidence is unavailable, Sketch ReID becomes essential. It focuses on matching hand-drawn sketches, often provided by eyewitnesses, with RGB surveillance images. This task is inherently challenging due to the large modality gap—sketches lack color, texture, and fine-grained detail, and they vary significantly in drawing style. Moreover, collecting large-scale annotated sketch datasets is costly, making generalization in real-world scenarios even more difficult. Existing Sketch ReID methods~\cite{pang2018cross,gong2021eliminate, ge2024robust,chen2024msif} primarily focus on aligning cross-modal features and learning modality-invariant embeddings. While effective in controlled benchmarks, they often rely on abundant labeled sketch data and struggle to generalize under limited supervision. This label scarcity, coupled with high inter-domain variation, calls for a learning framework that can transfer foundational visual knowledge to sketch domains using only a few examples.

To this end, we propose KTCAA, a meta-learning-based framework designed for few-shot cross-modal ReID. Inspired by recent advances in meta-learning~\cite{sung2018learning}, our method transfers knowledge from large-scale single-modal RGB data to address data-scarce cross-modal scenarios. The training process simulates low-shot tasks through meta-training and meta-testing phases, enabling the model to acquire transferable knowledge and adapt quickly to new domains with minimal sketch data. Critically, unlike previous works \cite{pang2018cross,gong2021eliminate,liu2024differentiable,lin2023beyond,zhu2022cross,ge2024robust,chen2024msif}, KTCAA is not merely a framework design, but is motivated by theoretical inspiration. We first revisit cross-modal transfer from the perspective of domain adaptation theory. By establishing a generalization error bound, we identify two necessary conditions for effective cross-modal transfer: domain discrepancy and perturbation invariance. These two conditions inform the design of our method.

\begin{figure*}[t]
	\centering
	\includegraphics[width=1\linewidth]{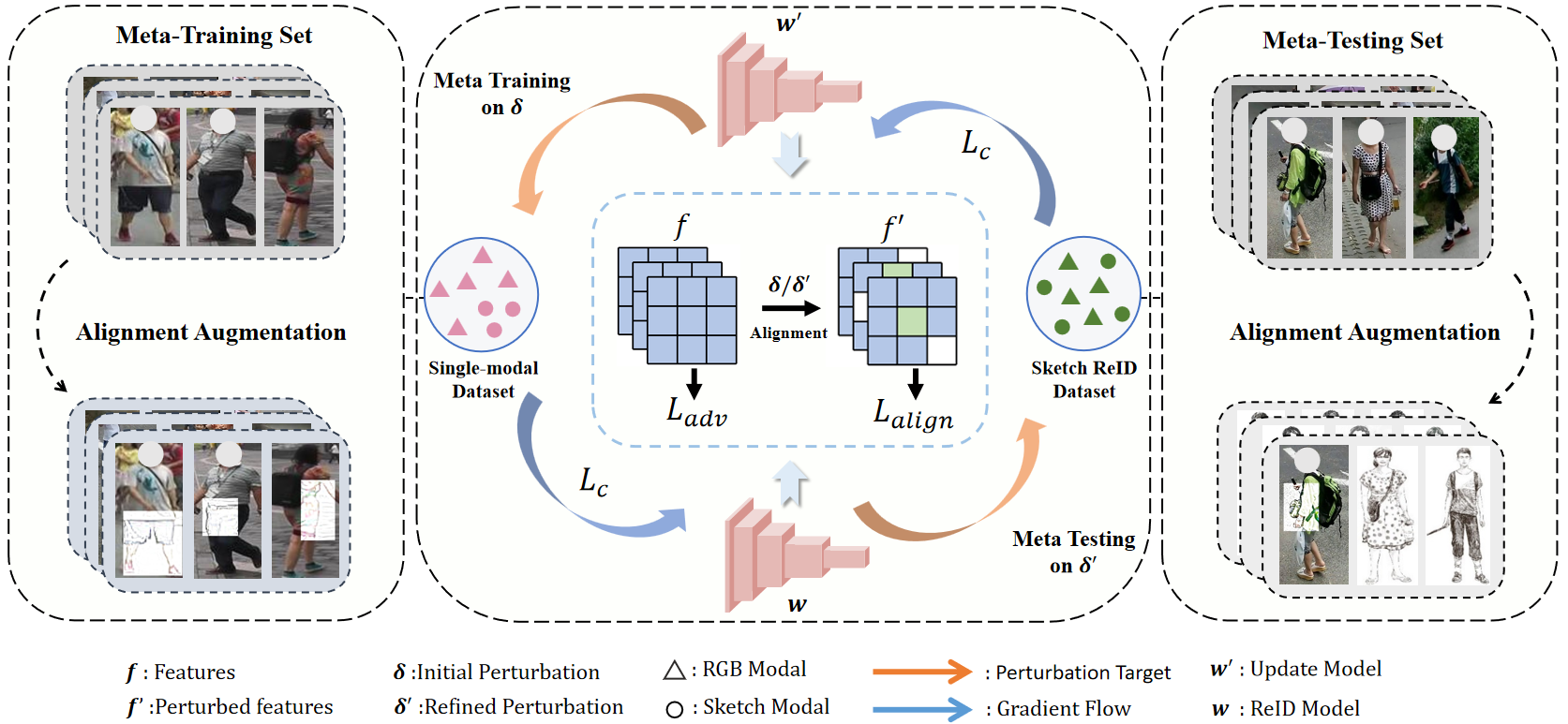}
	\caption{Pipeline of the proposed KTCAA framework, which operates under a meta-learning paradigm. During the meta-training phase, a single-modal RGB dataset is used. The Alignment Augmentation (AA) module applies localized sketch-style transformations to simulate target domain characteristics, modeling modality discrepancies at the image level. This guides the model to progressively align source and target distributions while preserving fine-grained semantics. The Knowledge Transfer Catalyst (KTC) module introduces adversarial perturbations to simulate cross-modal uncertainty and is jointly optimized through meta-learning. The alignment loss \(L_{\text{align}}\) between features before and after perturbation, along with the adversarial classification loss \(L_{\text{adv}}\), enhances the model’s robustness against detail blur and modality shifts. Additionally, the contrastive loss \(L_C\) is jointly optimized with these regularization terms to enhance cross-modal representation learning. During the meta-testing phase, the base model parameters \(w\) are frozen, and the updated model \(w'\) is fine-tuned for few-shot sketch-based Re-ID, leveraging cross-modal knowledge for improved generalization under domain shifts.}
	\label{aug}
\end{figure*}

To satisfy these conditions, we propose two theoretically inspired components. First, the Alignment Augmentation (AA) module reduces the distribution divergence between RGB and sketch modalities by performing local sketch-style transformations on RGB inputs. This localized augmentation simulates domain shifts while preserving identity structure. Second, the Knowledge Transfer Catalyst (KTC) module improves the model’s robustness to modality shifts by introducing worst-case perturbations that simulate cross-modal variations. These components are jointly optimized under a meta-learning scheme that promotes robust, domain-adaptive feature representations.


An overview of KTCAA is shown in Fig.1. The AA module guides the model to learn shared visual cues across distributions through partial modality transformation. During meta-training, the local differences introduced by AA are used to generate adversarial gradients in KTC, which distort features to increase uncertainty and encourage robust matching. This process transforms the destabilizing nature of adversarial perturbations into a constructive mechanism for learning transferable representations. By aligning representations across views and tasks, the model learns modality-consistent embeddings that generalize well to unseen sketches. In KTCAA, knowledge transfer is realized through adversarial training with shared perturbation priors across the meta-learning stages. Specifically, during meta-training on the RGB domain, the model is encouraged to learn perturbation-invariant representations via the KTC module. These adversarial priors, rather than being specific to a particular domain, encode general robustness to modality shifts. During meta-testing, this generalizable robustness is transferred to sketch-based tasks, enabling the model to adapt effectively under few-shot conditions. This mechanism allows KTCAA to transfer cross-modal knowledge implicitly. 

The main contributions of this paper are summarized as follows:

$\bullet$ A theoretical analysis is conducted to establish a generalization bound for cross-modal transfer, revealing domain discrepancy and perturbation invariance as two key factors influencing performance. Based on this theoretical insight, a principled framework is proposed, featuring two modules—Alignment Augmentation and Knowledge Transfer Catalyst—that are explicitly designed to target the identified objectives.

$\bullet$ We propose KTCAA, a progressive meta-learning framework that effectively adapts from RGB to sketch domains with limited data. Experiments demonstrate that our method achieves superior performance and generalization compared to state-of-the-art approaches.

\section{Related Works}
\subsection{Cross-modal Person Re-identification}
With the development of deep learning \cite{pu2025robust,gu2024entropy,sun2024cifarwarehouse,yang2024robust,zheng2025towards,yang2022robust,SPIDER_ICML25}, cross-modal recognition techniques have made significant progress, offering new possibilities for addressing identity recognition challenges in complex surveillance and law enforcement scenarios. Cross-modal ReID is a critical task in security \cite{1,gu2025dual} and surveillance applications, aiming to match pedestrian images captured across different devices and modalities. In scenarios where RGB images are unavailable, such as in low-light or nighttime conditions, alternative modalities like infrared are essential. Traditional single-modal ReID \cite{pang2025robust,pang2025joint,10419038,lu2025llava,sun2019dissecting} focuses exclusively on RGB images, whereas cross-modal ReID \cite{li2025video,yang2024shallow,yang2022learning,yang2023translation,li2025shape,li2025robust,zhang2024magic,adca} addresses the challenges of aligning features across multiple modalities. 

Sketch ReID~\cite{gong2021eliminate,pang2018cross,zhu2022cross,ge2024robust,chen2024msif,lin2023beyond} represents a unique and particularly challenging cross-modal task due to the substantial semantic gap between sketches and real-world images. Unlike RGB or infrared images, sketches lack critical details such as color and texture, and they are often subject to stylistic variations introduced by different artists. This introduces significant inter-modality discrepancies. To address these challenges, prior work has proposed various solutions. Pang et al. \cite{pang2018cross} applied cross-domain adversarial learning to identify domain-invariant features between sketches and RGB images. Subsequent research explored semantic feature alignment \cite{chen2023sketchtrans} to reduce inconsistencies between these modalities. Chen et al. \cite{liu2024differentiable} introduced Differentiable Auxiliary Learning (DAL), which generates auxiliary modalities using deep neural networks to bridge the gap between sketches and RGB images. However, the scarcity of large annotated sketch datasets remains a major limitation, often leading to overfitting and poor generalization across datasets. Additionally, SS-reID \cite{lin2023beyond} have explored the use of pretrained multimodal models like CLIP \cite{radford2021learning} to aid feature alignment across modalities. While this approach can improve alignment in certain scenarios, it typically relies on additional attribute labels within datasets, which limits its applicability in more generalized settings. Furthermore, the substantial modality gap may restrict the effectiveness of CLIP in achieving robust alignment.

\subsection{Meta-Learning}
Meta-learning has been developed to address the challenge of adapting to new tasks with limited training samples, thereby enhancing generalization across diverse scenarios \cite{zhang2023style, guo2020learning}. Existing meta-learning approaches are typically categorized into three groups: metric-based methods \cite{sung2018learning}, model-based methods \cite{santoro2016meta}, and optimization-based methods \cite{finn2017model}. Among these, MAML (Model-Agnostic Meta-Learning) \cite{finn2017model}, an optimization-based approach, has gained prominence for its ability to learn well-initialized weights that can quickly adapt to new tasks by simulating the learning process on meta-test tasks.

Building on the principles of meta-learning, we aim to address the challenges of cross-modal ReID by leveraging abundant single-modal data to inject shared knowledge into few-shot cross-modal tasks. Our goal is to enhance the generalization of sketch-based re-identification tasks, particularly in unseen domains. By progressively refining shared feature representations and reducing modality discrepancies across meta-training and meta-testing, we enable more efficient cross-modal transfer without requiring extensive supervision from the target modality.


\section{Theoretical Motivation}
Sketch ReID faces the fundamental challenge of generalizing from the RGB modality (source domain) to the sketch modality (target domain) under small-sample conditions. To address this, we investigate the problem from a generalization perspective and, drawing inspiration from domain adaptation theory, arrive at an extended formulation that highlights two key factors: domain discrepancy and perturbation sensitivity. This formulation provides theoretical inspiration for our framework design.

\subsection{Cross-Modal Generalization Bound}

Following domain adaptation theory~\cite{ben2010theory}, the expected target domain risk can be decomposed into three components: the expected source domain risk, a distribution discrepancy term, and an irreducible residual:
\begin{equation}
	\epsilon_T(h) \leq \epsilon_S(h) + \tfrac{1}{2} d_{\mathcal{H} \Delta \mathcal{H}}(\mathcal{D}_S, \mathcal{D}_T) + \lambda^*,
\end{equation}
where $\epsilon_T(h)$ and $\epsilon_S(h)$ denote the expected risks of a hypothesis $h$ on the target and source domains, respectively, $d_{\mathcal{H} \Delta \mathcal{H}}$ measures the discrepancy between the two domains, and $\lambda^*$ represents the minimum joint error of the ideal hypothesis, typically considered task-specific and fixed.

We assume that the feature extractor $f$ is $L$-Lipschitz continuous and exhibits local robustness to input perturbations $\delta$, such that:
\begin{equation}
	\mathbb{E}_{x \sim \mathcal{D}_S} \|f(x + \delta) - f(x)\| \leq \gamma.
\end{equation}
Under this assumption, the source risk can be approximated by the empirical risk plus a perturbation-sensitive regularization term:
\begin{equation}
	\epsilon_S(h) \approx \epsilon_S(f) \leq \widehat{\epsilon}_S(f) + L \cdot \gamma,
\end{equation}
where $\widehat{\epsilon}_S(f)$ denotes the empirical source risk, and $L \cdot \gamma$ reflects the worst-case impact of input perturbations.

Inspired by robustness principles such as Virtual Adversarial Training (VAT)~\cite{miyato2018virtual} and adversarial data augmentation \cite{volpi2018generalizing}, we introduce the following extended heuristic generalization bound to capture the key factors influencing cross-modal transfer:
\begin{equation}
	\epsilon_T(h) \lesssim \widehat{\epsilon}_S(f) + L \cdot \gamma + \tfrac{1}{2} d_{\mathcal{H} \Delta \mathcal{H}} + \lambda^*.
	\label{eq:heuristic-bound}
\end{equation}
The term $L \cdot \gamma$ accounts for the model's sensitivity to input variations and provides a useful regularization perspective inspired by adversarial robustness. Although it is not part of the original bound proposed by Ben-David et al.~\cite{ben2010theory}, we adopt it as a conceptual objective to guide the design of our framework.

In this formulation, $d_{\mathcal{H} \Delta \mathcal{H}}(\mathcal{D}_S, \mathcal{D}_T)$ quantifies the discrepancy between source and target domains, reflecting how differently classifiers behave across modalities. To reduce this, the AA module locally transforms RGB images into sketch-like variants, aligning distributions at the input level and narrowing the domain gap.

Meanwhile, $\gamma$ captures the model’s sensitivity to input perturbations such as cross-modal shifts. Unlike $d_{\mathcal{H} \Delta \mathcal{H}}$, which focuses on data-level differences, $\gamma$ characterizes the stability of the feature extractor itself. To suppress it, the KTC module applies worst-case perturbations and enforces consistency through adversarial and alignment losses, thereby promoting robust and transferable representations.

\subsection{AA: Reducing Domain Discrepancy}

To reduce the domain divergence term $d_{\mathcal{H}\Delta\mathcal{H}}$, we introduce the Alignment Augmentation (AA) module, which simulates sketch-style variations by locally replacing regions in RGB images with their sketch-transformed counterparts. This localized perturbation preserves global structure while introducing target-modality characteristics.

Formally, given an input image $x$ and its sketch-transformed version $\Delta_m(x)$, we define:
\begin{equation}
	\delta_{\text{global}} = \Delta_m(x) - x, \quad \delta_{\text{local}} = M \odot (\Delta_m(x) - x),
\end{equation}
where $M \in \{0,1\}^{H \times W}$ is a binary mask selecting the perturbed region, and $\odot$ denotes element-wise multiplication.

Assuming the feature extractor $f$ is $L$-Lipschitz, we have:
\begin{equation}
	\|f(x + \delta) - f(x)\| \leq L \cdot \|\delta\|.
\end{equation}
Since $\|\delta_{\text{local}}\| < \|\delta_{\text{global}}\|$, the resulting feature perturbation is smaller under local augmentation. This allows the model to preserve semantic structure while introducing modality-specific features more smoothly, thereby facilitating cross-domain alignment and effectively reducing $d_{\mathcal{H}\Delta\mathcal{H}}$.

\subsection{KTC: Minimizing Perturbation Sensitivity}
To suppress the perturbation response $\gamma$, we adopt a robustness-inspired training objective based on worst-case input shifts. Inspired by VAT, we approximate the expected perturbation sensitivity via an adversarial upper bound:
\begin{equation}
	\gamma = \mathbb{E}_x \|f(x + \delta) - f(x)\| \lesssim \max_{\|\eta\| \leq \epsilon} \|f(x + \eta) - f(x)\|.
\end{equation}
Here, the left-hand side represents the model’s average sensitivity to random input perturbations $\delta$, while the right-hand side approximates the worst-case response under bounded adversarial perturbations $\eta$ satisfying $\|\eta\| \leq \epsilon$.
This upper bound serves as a tractable surrogate that captures the model’s robustness to modality shifts. To implement this, the Knowledge Transfer Catalyst (KTC) module generates adversarial sketch-style perturbations and enforces feature consistency via the alignment loss:
\begin{equation}
	\mathcal{L}_{\text{align}} = \|f(x + \eta) - f(x) \|_2^2.
\end{equation}
This encourages the model to learn perturbation-invariant representations and effectively reduces $\gamma$, supporting better cross-modal generalization.

\section{Methodology}

Motivated by the theoretical analysis of the generalization bound in cross-modal learning, we identify two key factors for robust transfer: domain discrepancy and perturbation stability. These insights guide the design of our proposed framework, KTCAA, which adopts a dual-path architecture based on ResNet50. It extracts modality-specific features for RGB and sketch images through independent shallow branches, while sharing upper fully connected layers for unified representation. The framework consists of two theory-inspired modules: Alignment Augmentation (AA), which reduces domain discrepancy and facilitates progressive alignment through localized modality transformations that simulate target sketch-style characteristics; and Knowledge Transfer Catalyst (KTC), which improves stability under latent modality variations by enforcing consistency under adversarial modality perturbations. These two modules are integrated into a meta-learning strategy designed to improve adaptation in cross-domain few-shot scenarios through joint optimization.


\subsection{Meta-Training}
\textbf{Alignment Augmentation.}
The process of alignment augmentation can be represented as follows:
\begin{equation}
	x_{i}^{sketch} = t(x_{i}^{rgb}),
\end{equation}
where \( t(\cdot) \) denotes the function that converts an RGB image into a sketch image. Furthermore, if it is necessary to select a local region at a random position in the image for sketching, this process can be expressed as:
\begin{equation}
	x_{i}^{ls} = LS(x_{i}^{rgb}, x_{i}^{sketch}, rect),
\end{equation}
where \( rect \) represents the randomly selected rectangular area, and \( LS(\cdot) \) indicates the operation of replacing the local region in the RGB image with the corresponding sketch. More details can be found in the Appendix.

We use InfoNCE loss as the contrastive loss, which is defined as:
\begin{equation}
	L_{c} = -\frac{1}{N} \sum_{i=1}^{N} \log \frac{\exp(\text{sim}(\mathbf{z}_i, \mathbf{z}_i^+) / \tau)}{\sum_{j=1}^{N} \exp(\text{sim}(\mathbf{z}_i, \mathbf{z}_j) / \tau)},
\end{equation}
where \( \mathbf{z}_i \) denotes the feature representation of the original RGB image, and \( \mathbf{z}_i^+ \) represents the feature representation of either the alignment augmentation sample or a sketch sample of the same identity. Here, \( \mathbf{z}_j \) refers to the representation of negative samples with different identities from \( \mathbf{z}_i \), \( \text{sim}(\cdot) \) is a similarity function, \( \tau \) is a temperature parameter that controls the smoothness of the distribution (set to \( \tau=0.1 \)), and \( N \) is the batch size. This loss function encourages the model to learn discriminative features by maximizing the similarity between positive pairs (\( \mathbf{z}_i \) and \( \mathbf{z}_i^+ \)) while minimizing the similarity between negative pairs (\( \mathbf{z}_i \) and \( \mathbf{z}_j \)).

\textbf{Knowledge Transfer Catalyst.}
The goal of the Knowledge Transfer Catalyst is to optimize knowledge transfer by learning universal perturbations~\cite{moosavi2017universal,1}. By maximizing potential differences between modalities, these perturbations encourage the model to learn robust feature representations. In the meta-learning and meta-training phases, the information interactions facilitated by these universal perturbations expose the model to diverse cross-modal variations, thereby improving the stability of feature learning and enhancing the effectiveness of knowledge transfer.

We design triplet loss to distort the pairwise relationships between pedestrian identities. For a specific batch of \( n \) samples that includes both original and alignment augmentation images, we utilize the model to extract and perturb their features. The temporary perturbation \( \eta \) is iteratively updated using Momentum Inertia Stochastic Gradient Descent (MI-SGD), as shown below:
\begin{align}
	L_{RGB}(f^{adv}_{RGB}) = \max \big[ \ 
	& \lVert f^n_{RGB} - f^{adv}_{RGB} \rVert_2 \notag \\
	& - \lVert f^p_{RGB} - f^{adv}_{RGB} \rVert_2 + \rho, \ 0 \ 
	\big], \label{f5}
\end{align}
\( \rho \) is a margin parameter that controls the minimum distance between the negative pair and the positive pair. It can be used to bring negative pairs closer together and push positive pairs further apart.
\begin{equation}
	\begin{aligned}
		\Delta_{meta\_train} = \theta \Delta'_{meta\_test} + \frac{\nabla_{\eta}L_{RGB}}{\| \nabla_{\eta}L_{RGB} \|_1 },
	\end{aligned}
	\label{loss1}
\end{equation}
\begin{equation}
	\begin{aligned}
		\eta = \text{clip}(\eta + \alpha \cdot \text{sign}(\Delta_{meta\_train}), -\varepsilon,\varepsilon).
	\end{aligned}
	\label{loss2}
\end{equation}
Here, \( \theta \) represents the momentum value, \( \Delta'_{meta\_test} \) is derived from the previous iteration, \( \alpha \) denotes the iteration step size, and \( \varepsilon \) is the adversarial bound. The above is the optimization process for the universal perturbation \( \eta \), which results in the adversarial sample:
\begin{equation}
	x_{adv} = x + \eta .
\end{equation}
Using adversarial samples for adversarial training, the adversarial loss is:
\begin{equation}
	L_{adv}(\theta) = -\mathbb{E}_{x \sim \mathcal{D}} \left[ \sum_{i=1}^{C} y_i \log(f(x + \eta)) \right],
\end{equation}
where \( C \) is the number of classes, \( y_i \) corresponds to the true label, \( \eta \) is the generated adversarial perturbation.

To ensure consistency between features before and after perturbation, we define the alignment loss as follows:
\begin{equation}
	L_{\text{align}} = \| f(x) - f(x_{adv}) \|_2^2,
\end{equation}
where \( f(x) \) represents the feature vector of the original sample \( x \), and \( f(x_{adv}) \) represents the feature vector of the perturbed sample \( x_{adv} \). This loss term enforces the model to produce similar feature representations for the original and perturbed samples.

The overall loss function for the meta-training phase, denoted as \( L_{meta\_train} \):
\begin{equation}
	L_{meta\_train} = L_c + L_{adv} + L_{\text{align}}.
\end{equation}
Here, \( L_c \) promotes discriminative feature learning via contrastive learning, \( L_{adv} \) encourages robust feature learning, and \( L_{\text{align}} \) ensures feature consistency between original and perturbed samples. The combination of these losses enables robust cross-modal alignment and generalization.

\subsection{Meta-Testing}
This step is similar to meta-training. We utilize the sketch ReID dataset to learn the perturbation $\eta$ with the our loss functions:
\begin{align}
	L_{s}(f^{adv}_{RGB}) =\ & \max \Big[ \|f^n_{s} - f^{adv}_{RGB}\|_2 \nonumber \\
	& - \|f^p_{s} - f^{adv}_{RGB}\|_2 + \rho,\ 0 \Big],
	\label{f9}
\end{align}
where \( f^n_{s} \),\( f^p_{s} \) and \( f^{adv}_{RGB} \) denote the feature vectors of the sketch image and the RGB adversarial sample in the sketch ReID dataset, respectively.
\begin{equation}
	\begin{aligned}
		\Delta_{meta\_test} = \theta \Delta_{meta\_train} + \frac{\nabla_{\eta}L_{s}}{\| \nabla_{\eta}L_{s} \|_1 },
	\end{aligned}
	\label{loss3}
\end{equation}
\begin{equation}
	\begin{aligned}
		\eta = \text{clip}(\eta +	 \alpha \cdot \text{sign}(\Delta_{meta\_test}), -\varepsilon,\varepsilon).
	\end{aligned}
	\label{loss4}
\end{equation}
Here, $\Delta_{meta\_train}$ derived from step of meta-training. The main difference compared to the previous step lies in the perturbation applied to the input and the gradients related to momentum. 

Similarly, the adversarial loss \( L_{adv} \) in the meta-testing phase can be obtained from equations (8) and (9). The alignment loss \( L_{\text{align}} \) can then be derived from equation (10). Therefore, the loss in the meta-testing phase can be expressed as follows:
\begin{equation}
	L_{meta\_test}= L_c + L_{adv} + L_{\text{align}}.
\end{equation}

\subsection{Meta-Update}
The final step is based on the losses from the previous two steps. Specifically, our ultimate loss function can be expressed as:
\begin{equation}
	L_{total} = L_{meta\_train} + \ L_{meta\_test}.
\end{equation}

The former aims to learn foundational knowledge through meta-training to build the model's cross-modal feature representation capability, while the latter seeks to effectively transfer this acquired knowledge to the sketch ReID task.

\begin{table*}[t]
	\small
	\centering
	\begin{minipage}[t]{0.48\linewidth}
		\setlength{\tabcolsep}{3.5pt}
		\centering
		\textbf{(a) Market-Sketch-1K Dataset}\\
		\begin{tabular}{l c c}
			\toprule
			\textbf{Methods} & \textbf{Rank-1 (\%)} & \textbf{mAP (\%)} \\ 
			\midrule
			DSCNet~\cite{zhang2022dual}         & 13.8  & 14.7 \\ 
			DEEN~\cite{zhang2023diverse}        & 12.1  & 12.6 \\
			UNIReID~\cite{chen2023towards}      & 14.6  & 13.2 \\ 
			AIO~\cite{li2024all}                & 15.4  & 13.9 \\ 
			SS-reID~\cite{lin2023beyond}        & 18.1  & 19.6 \\  
			\textbf{Ours}                       & \textbf{21.3} & \textbf{22.7} \\ 
			\bottomrule
		\end{tabular}
		\label{t2}
	\end{minipage}
	\hfill
	\begin{minipage}[t]{0.48\linewidth}
		\setlength{\tabcolsep}{3.5pt}
		\centering
		\textbf{(b) PKU-Sketch Dataset}\\
		\begin{tabular}{l c c}
			\toprule
			\textbf{Methods} & \textbf{Rank-1 (\%)} & \textbf{mAP (\%)} \\ 
			\midrule
			AFLNet~\cite{pang2018cross}               & 34.0  & -    \\ 
			CCSC~\cite{zhang2022cross}               & 86.0  & 83.7 \\ 
			SS-reID~\cite{lin2023beyond}             & 78.0  & 78.7 \\  
			DALNet~\cite{liu2024differentiable}      & 90.0  & 86.2 \\ 
			UNIReID~\cite{chen2023towards}           & 91.4  & 91.8 \\ 
			\textbf{Ours}                            & \textbf{92.7} &   \textbf{92.3} \\ 
			\bottomrule
		\end{tabular}
		\label{t1}
	\end{minipage}
	\caption{Comparison with SOTA methods. Left: Market-Sketch-1K (single-query); Right: PKU-Sketch. Market1501 is not used as auxiliary data to ensure fair comparison.}
	\label{tab:combined_comparison}
\end{table*}

\begin{table*}[ht]\small
	\centering
	\begin{tabular}{ll cc cc cc}
		\toprule
		\textbf{Dataset} & \textbf{Method} & \multicolumn{2}{c}{\textbf{80\% data}} & \multicolumn{2}{c}{\textbf{60\% data}} & \multicolumn{2}{c}{\textbf{40\% data}} \\ 
		\cmidrule(r){3-4} \cmidrule(r){5-6} \cmidrule(r){7-8}
		& & \textbf{Rank-1 (\%)} & \textbf{mAP (\%)} & \textbf{Rank-1 (\%)} & \textbf{mAP (\%)} & \textbf{Rank-1 (\%)} & \textbf{mAP (\%)} \\
		\midrule
		\multirow{2}{*}{\textbf{PKU-Sketch}} 
		& UNIReID & 83.3 ± 0.8 & 78.0 ± 0.3 & 69.6 ± 0.2 & 64.3 ± 0.7 & 44.0 ± 0.7 & 40.1 ± 0.5 \\
		& \textbf{Ours} & \textbf{90.0 ± 0.2} & \textbf{85.0 ± 0.4} & \textbf{85.1 ± 0.3} & \textbf{80.3 ± 0.5} & \textbf{68.2 ± 0.3} & \textbf{64.1 ± 0.2} \\
		\midrule
		\multirow{2}{*}{\textbf{Market-Sketch-1k}} 
		& SS-reID & 12.2 ± 0.4 & 11.0 ± 0.3 & 9.8 ± 0.5 & 9.2 ± 0.4 & 6.5 ± 0.6 & 6.0 ± 0.4 \\
		& \textbf{Ours} & \textbf{21.5 ± 0.3} & \textbf{23.1 ± 0.3} & \textbf{18.2 ± 0.4} & \textbf{19.7 ± 0.3} & \textbf{14.8 ± 0.5} & \textbf{16.3 ± 0.4} \\
		\bottomrule
	\end{tabular}
	\caption{Comparison of cross-modal performance under limited data. Experiments are conducted on the PKU-Sketch and Market-Sketch-1K datasets, with the latter evaluated under the single-query setting. Each episode adopts a 5-way setup by sampling 5 sketch identities and using their corresponding RGB samples as support. Meta-test identities are disjoint from meta-train to ensure fair evaluation.}
	\label{data_scarcity_combined}
\end{table*}

\begin{table*}[h]\small
	\centering
	\begin{tabular}{c c c c c c cc cc}
		\toprule
		\textbf{Baseline} & \textbf{+AA} & \textbf{+KTC} & \textbf{+$\mathbf{L_{c}}$} & \textbf{+$\mathbf{L_{\text{adv}}}$} & \textbf{+$\mathbf{L_{\text{align}}}$} & \multicolumn{2}{c}{\textbf{Market-sketch-1K}} & \multicolumn{2}{c}{\textbf{PKU-Sketch}} \\ 
		&              &               &                   &                                &                              & \textbf{Rank-1 (\%)} & \textbf{mAP (\%)}        & \textbf{Rank-1 (\%)} & \textbf{mAP (\%)} \\ 
		\midrule
		\checkmark &              &               &                   &                                &                              & 12.1        & 15.3                & 68.2        & 69.0        \\ 
		\checkmark &              & \checkmark    & \checkmark        & \checkmark                     &                    & 17.4        & 18.7                & 76.5        & 75.1        \\ 
		\checkmark & \checkmark   &               & \checkmark        &                                &                              & 18.0        & 19.9                & 80.9        & 80.2        \\ 
		\checkmark & \checkmark   & \checkmark    & \checkmark        & \checkmark                     &                              & 23.2        & 23.5                & 87.9        & 86.4        \\ 
		\checkmark & \checkmark   & \checkmark    & \checkmark        & \checkmark                     & \checkmark                   & 24.9        & 26.2                & 93.6        & 92.9        \\ 
		\bottomrule
	\end{tabular}
	\caption{Evaluation of each component of the proposed method on the Market-Sketch-1K and PKU-Sketch datasets.}
	\label{component_evaluation}
\end{table*}

\section{Experiments}

\subsection{Experimental Settings}
We conduct experiments on three benchmark datasets. PKU-Sketch~\cite{pang2018cross} contains hand-drawn sketches and corresponding RGB images of pedestrians under diverse poses and backgrounds, targeting sketch-based cross-modal ReID. Market-Sketch-1K~\cite{lin2023beyond} introduces artist subjectivity to simulate subtle sketch variations, offering a realistic testbed for generalization under modality and style discrepancies. Market-1501~\cite{zheng2015scalable} is a large-scale RGB-only dataset containing 1,501 identities captured from six camera views. In this work, we use it as a single-modal source dataset for transfer learning. Unless otherwise specified, all experiments are conducted with Market-1501 as the default auxiliary training set to support cross-modal transfer learning.

For evaluation, we follow standard ReID protocols~\cite{pang2018cross,zhang2022cross}, reporting Rank-1 accuracy and mean Average Precision (mAP). Rank-1 measures the retrieval accuracy of the top-1 result, while mAP reflects the overall ranking quality of the retrieved results.

Our experiments are conducted on 2 NVIDIA A40 GPUs. In our KTC module setup, we configure the hyperparameters as follows: we set the maximum iteration count \( \text{max\_iter} = 10 \), with a margin \( \rho = 0.5 \) and a learning rate \( \alpha = \epsilon / 10 \). We use stochastic gradient descent (SGD) with a momentum factor \( \theta = 0.9 \) to update \( \eta \). For \( L_{\infty} \) bounded attacks, \( \epsilon \) is set to 8. Given the imbalance in dataset sizes between meta-training and meta-testing, we conduct one meta-test after every 10 meta-training iterations, with a batch size \( N_b = 32 \).

\subsection{Comparison with State-of-the-Art Methods}
Tab.\ref{tab:combined_comparison} presents a comparison of our method with SOTA approaches on the Market-Sketch-1K and PKU-Sketch datasets. For fairness, no single-modal data is used in this evaluation. On Market-Sketch-1K, our method achieves 21.3\% Rank-1 and 22.7\% mAP, significantly outperforming SS-reID~\cite{lin2023beyond} (18.1\% Rank-1, 19.6\% mAP). This gain stems from the proposed AA and KTC modules, which improve cross-modal alignment and robustness, especially in handling subjective variations from different sketch artists. On PKU-Sketch, our method reaches 92.7\% Rank-1 and 92.3\% mAP, surpassing UNIReID (91.4\% Rank-1, 91.8\% mAP). While DALNet improves sketch-photo alignment via auxiliary modalities, our method explicitly tackles the two key factors in domain adaptation theory—domain discrepancy and perturbation stability—through alignment augmentation and adversarial perturbations. This enables our model to generalize effectively across datasets. In the Appendix, we further evaluate the model’s generalization to unseen domains via cross-dataset testing between PKU-Sketch and Market-Sketch-1K. When trained on PKU and tested on Market-Sketch-1K, our method achieves 15.4\% Rank-1 and 12.7\% mAP, outperforming SS-reID (11.2\% / 10.6\%). In the reverse direction, our method reaches 38.1\% Rank-1 and 32.6\% mAP vs. SS-reID’s 32.8\% / 29.2\%. These results demonstrate that KTCAA exhibits stronger cross-domain generalization even without auxiliary labels, reinforcing its effectiveness under few-shot and domain-shift scenarios.

\subsection{Transfer Learning Under Data Scarcity}

Tab.\ref{data_scarcity_combined} reports the performance of KTCAA under varying data availability on the PKU-Sketch and Market-Sketch-1K datasets. All models are pre-trained on Market-1501 and fine-tuned using 80\%, 60\%, and 40\% of the training data. Results are reported as the mean ± standard deviation over multiple runs. On PKU-Sketch, KTCAA consistently outperforms UNIReID across all data scales. For example, at the 40\% level, KTCAA achieves substantially higher Rank-1 accuracy and mAP compared to UNIReID. On the more challenging Market-Sketch-1K, our method also shows clear advantages over SS-reID in all settings. These results demonstrate KTCAA’s superior data efficiency and cross-modal generalization under few-shot conditions.

\subsection{Ablation Experiment}
Tab.\ref{component_evaluation} presents the ablation results on Market-Sketch-1K and PKU-Sketch. Starting from a baseline, we progressively introduce the AA and KTC modules along with their corresponding losses. On Market-Sketch-1K, the AA module injects localized modality transformations to guide progressive distribution alignment, improving Rank-1 accuracy from 12.1\% to 18.0\%. Introducing the KTC module (with adversarial perturbation and $L_{\text{adv}}$ loss) further improves robustness to modality shifts, boosting Rank-1 to 23.2\%. The full model, with both KTC and the alignment loss $L_{\text{align}}$, achieves the highest performance with 24.9\% Rank-1 and 26.2\% mAP, demonstrating the complementary benefits of AA and KTC for alignment and generalization. To better understand the mechanism behind these gains, we conduct additional experiments in the Appendix. First, we compare global versus local augmentation and show that the \textit{AA} module achieves better generalization than a global sketch-style transformation. Next, we perform a sensitivity analysis on two key KTC hyperparameters: the adversarial bound $\varepsilon$ and the triplet margin $\rho$. Results indicate that $\varepsilon = 8$ and $\rho = 0.5$ provide the best trade-off between performance and efficiency. Finally, we report training times under different settings. While KTC increases computational cost due to adversarial optimization, it does not affect inference time, and the performance gains justify the additional training effort.

\begin{table}[t]
	\centering
	\small
	\setlength{\tabcolsep}{3pt}
	\begin{tabular}{ccccc}
		\toprule
		\textbf{Baseline} & \textbf{+CLIP-Align} & \textbf{+KTCAA} & \textbf{Rank-1 (\%)} & \textbf{mAP (\%)} \\
		\midrule
		\checkmark &          &          & 16.9 & 17.4 \\
		\checkmark & \checkmark &        & 21.4 & 20.8 \\
		\checkmark &          & \checkmark & 24.7 & 25.2 \\
		\checkmark & \checkmark & \checkmark & \textbf{25.1} & \textbf{26.5} \\
		\bottomrule
	\end{tabular}
	\caption{Comparison of cross-modal alignment performance based on the SS-reID baseline on the Market-Sketch-1K dataset.}
	\label{alignment}
\end{table}

\begin{figure}[t]
	\centering
	\includegraphics[width=\linewidth]{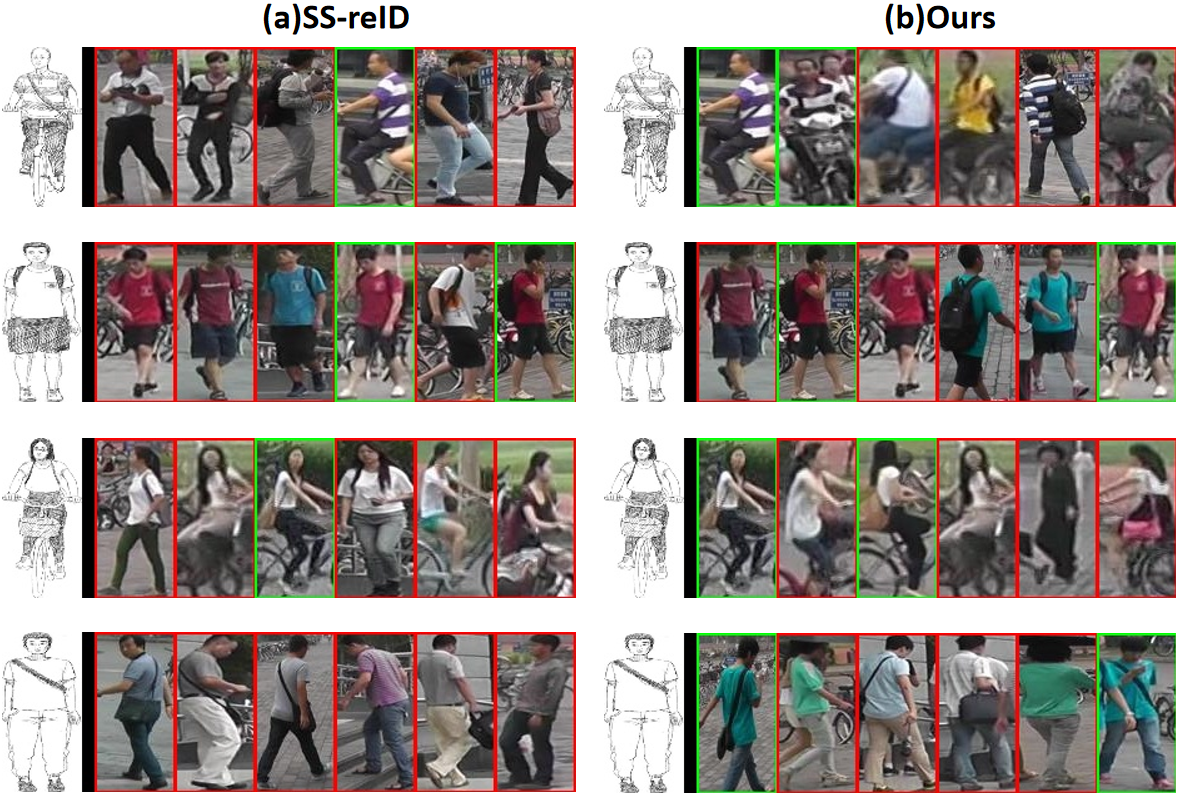}
	\caption{
		Qualitative comparison of retrieval results between SS-reID (a) and our KTCAA (b) on the Market-Sketch-1K dataset. Each row shows top-6 results for a sketch query. Green and red boxes indicate correct and incorrect matches, respectively. 
	}
	\label{qualitative_results} 
\end{figure}

\subsection{Comparison of Cross-Modal Alignment}  
On the Market-Sketch-1K dataset, we evaluate the performance of the SS-reID baseline model and its integration with the CLIP alignment module (CLIP-Align). For a fair comparison, the competing methods were pre-trained on the Market1501 dataset. 

As shown in Tab.\ref{alignment}, the SS-reID baseline achieves a Rank-1 accuracy of {16.9\%} and an mAP of {17.4\%}, highlighting its limitations in cross-modal alignment. Incorporating the CLIP-Align module improves the Rank-1 accuracy to {21.4\%} and mAP to {20.8\%}. With our proposed KTCAA framework, these metrics are further boosted to {25.1\%} Rank-1 and {26.5\%} mAP, achieving the highest performance. This result highlights KTCAA’s effectiveness in simultaneously reducing domain discrepancy and improving perturbation stability, thereby achieving stronger cross-modal alignment and generalization, even when integrated with existing alignment methods.

\subsection{Visualization}

To qualitatively assess the effectiveness of KTCAA, we visualize top-6 retrieval results in Fig.\ref{qualitative_results}. Compared to SS-reID, KTCAA retrieves more accurate matches with better consistency across pose, viewpoint, and modality variations, demonstrating stronger cross-modal alignment and discrimination. In addition, t-SNE visualizations provided in the Appendix show that KTCAA yields more compact identity clusters and better separation across modalities, further validating its alignment capability at the feature level.

\section{Conclusion}
This work introduces KTCAA, a theoretically inspired framework for few-shot sketch-based person re-identification. Inspired by generalization theory, KTCAA addresses domain discrepancy and perturbation sensitivity via two modules: AA, which injects sketch-style cues for cross-modal alignment, and KTC, which enhances robustness by reducing worst-case feature shifts. Jointly trained under a meta-learning paradigm, KTCAA enables effective RGB-to-sketch transfer and achieves state-of-the-art performance in low-data scenarios.

\section*{Acknowledgment}
This work was supported in part by the National Natural Science Foundation of China under Grant No. 62276222.

\bibliographystyle{unsrt}
\bibliography{main}

\end{document}